\pdfoutput=1

\documentclass[11pt]{article}

\usepackage[preprint]{acl}

\usepackage{times}
\usepackage{latexsym}

\usepackage[T1]{fontenc}

\usepackage[utf8]{inputenc}

\usepackage{microtype}

\usepackage{inconsolata}

\usepackage{graphicx}

\usepackage{booktabs}
\usepackage{multirow}
\usepackage{amsmath}
\usepackage{subcaption}
\usepackage{colortbl}  
\usepackage{amsfonts}

\newcommand{\ours}{\texttt{RoleRAG}}
\newcommand{\ie}{\textit{i.e.}}
\newcommand{\eg}{\textit{e.g.}}

\newif\ifrevision
\revisionfalse
\ifrevision
\newcommand{\revise}[1]{\textcolor[RGB]{68,114,196}{#1}}
\else
\newcommand{\revise}[1]{#1}
\fi

\title{\textit{Single LLM, Multiple Roles}: A Unified Retrieval-Augmented Generation Framework Using Role-Specific Token Optimization}

\author{Yutao Zhu$^1$, Jiajie Jin$^1$, Hongjin Qian$^{2}$, Zheng Liu$^{2}$, Zhicheng Dou$^{1}$\thanks{Corresponding author.}\and Ji-Rong Wen$^1$ \\
$^1$Gaoling School of Artificial Intelligence, Renmin University of China \\
$^2$Beijing Academy of Artificial Intelligence, China \\
\texttt{yutaozhu94@gmail.com, dou@ruc.edu.cn}
}

\begin{document}
\maketitle
\begin{abstract}
Existing studies have optimized retrieval-augmented generation (RAG) across various sub-tasks, such as query understanding and retrieval refinement, but integrating these optimizations into a unified framework remains challenging. To tackle this problem, this work proposes \ours{}, a unified RAG framework that achieves efficient multi-task processing through role-specific token optimization. \ours{} comprises six modules, each handling a specific sub-task within the RAG process. Additionally, we introduce a query graph to represent the decomposition of the query, which can be dynamically resolved according to the decomposing state. All modules are driven by the same underlying LLM, distinguished by task-specific role tokens that are individually optimized. This design allows \ours{} to dynamically activate different modules within a single LLM instance, thereby streamlining deployment and reducing resource consumption. Experimental results on five open-domain question-answering datasets demonstrate the effectiveness, generalizability, and flexibility of our framework.
\end{abstract}

\section{Introduction}
Large language models (LLMs) have demonstrated remarkable performance across a wide range of tasks~\cite{gpt-3,gpt-4,llama3,deepseek_v3}. While their super power is driven by extensive parameters and large-scale training data, they still face challenges related to accuracy, reliability, and timeliness. Retrieval-augmented generation (RAG) provides an effective solution to these problems~\cite{sure,selfrag,rqrag}. By integrating an external retriever, the LLMs can access relevant knowledge based on user input queries, thus producing more accurate and reliable responses. This approach is particularly beneficial for knowledge-intensive tasks, such as open-domain question answering~\cite{kilt}.

\begin{figure}[t]
    \centering
    \includegraphics[width=.9\linewidth]{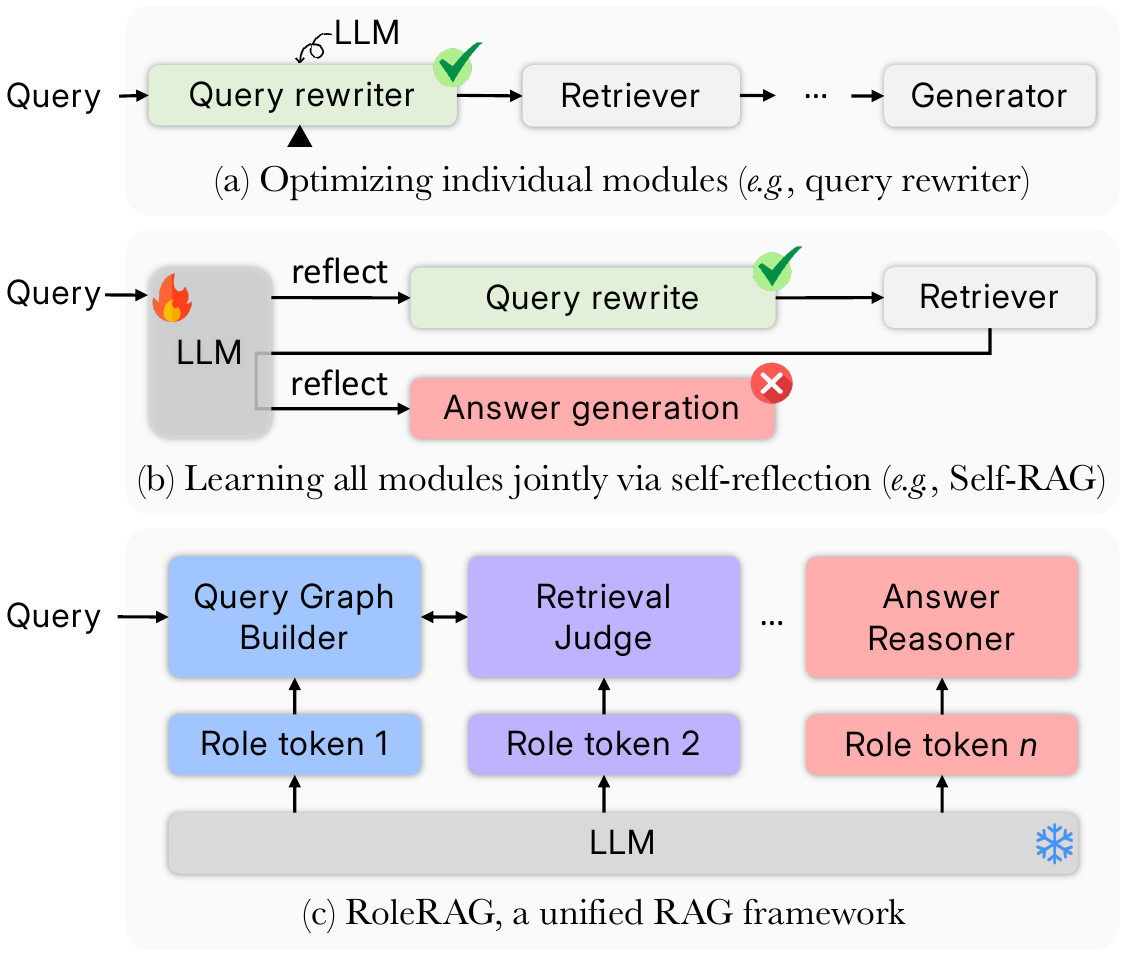}
    \caption{Comparison between existing studies and our framework.}
    \label{fig:intro}
    \vspace{-10px}
\end{figure}

\revise{In general, existing studies on optimizing RAG framework can be roughly categorized into two groups.} The first group focuses on \textbf{improving specific modules} of the framework, as shown in Figure~\ref{fig:intro}~(a). These efforts include introducing a retrieval necessity judgment module to reduce retrieval costs~\cite{slimplm,DBLP:journals/corr/abs-2411-06207}, improving query understanding to construct more effective queries for retrieving relevant knowledge~\cite{ircot,rqrag}, and refining retrieved results to extract key information that helps LLMs generate more accurate responses~\cite{sure,bider}. These enhancements have demonstrated improvements in the overall performance of RAG systems. However, integrating these diverse optimizations into a single unified framework is non-trivial. \revise{The second group attempts to \textbf{consolidate multiple RAG components within a single LLM}, utilizing self-reflection mechanisms to dynamically control the response generation process~\cite{selfrag,rqrag} (Figure~\ref{fig:intro}~(b)).} While this approach offers a straightforward framework, it faces two main challenges: (1) introducing new components, such as query rewriter, requires additional data collection and extensive re-training of the LLM, and (2) incorporating more functions increases the complexity of the model's self-reflection mechanism, which may degrade both performance and generalization capabilities.

To address these challenges, we propose \textbf{\ours}, \textbf{a unified RAG framework} using role-specific token optimization (illustrated in Figure~\ref{fig:intro}~(c)). \ours{} comprises six specialized modules: query graph builder, retrieval judge, sub-answer generation, summarizer, new query generator, and answer reasoner. The workflow begins with the query graph builder, which decomposes the input query into multiple sub-queries and constructs a directed acyclic graph. For each sub-query, the retrieval judge determines whether additional knowledge retrieval is necessary or if it can be answered directly. Based on this decision, the sub-answer generator produces a response. For sub-queries requiring external knowledge, the summarizer extracts key information from the retrieved content and updates an answer memory dictionary that stores the sub-query, retrieved data, and generated answer. Once all sub-queries are processed, the new query generator examines the answer memory and the original query to determine if further sub-queries are needed. Finally, the answer reasoner synthesizes the final response. 

\ours{} employs a \textbf{role-specific token optimization strategy} to implement these modules. By introducing additional special tokens and optimizing their embeddings using task-specific data, the framework enables each module to perform its designated function effectively.  Importantly, only the role tokens are tuned during training, making the training computationally efficient. During inference, a single LLM instance is deployed, with different role tokens acting as soft prompts to dynamically activate the corresponding modules. These modules collaborate in an iterative manner to generate the final response. Experimental results on five open-domain question-answering datasets demonstrate that \ours{} achieves performance improvements of 16\%-64\% over state-of-the-art RAG methods in terms of exact match score. Additional experiments further confirm the generalizability and robustness of our framework.

Our contributions are three-fold:

(1) We introduce a unified RAG framework using a role-specific token optimization strategy. By integrating a frozen backbone LLM with adaptive role tokens, the model can specialize in different modules of the RAG pipeline and collaborate effectively to complete the full process.

(2) We propose a query graph construction approach to improve the handling of complex queries in RAG. Our framework dynamically refines the query graph by eliminating redundant sub-queries and generating new ones when necessary, enhancing retrieval efficiency and relevance.

(3) We release a comprehensive dataset to train different RAG modules. To our best knowledge, this is the first dataset covering the entire pipeline of a RAG system. 

\begin{figure*}
    \centering
    \includegraphics[width=\linewidth]{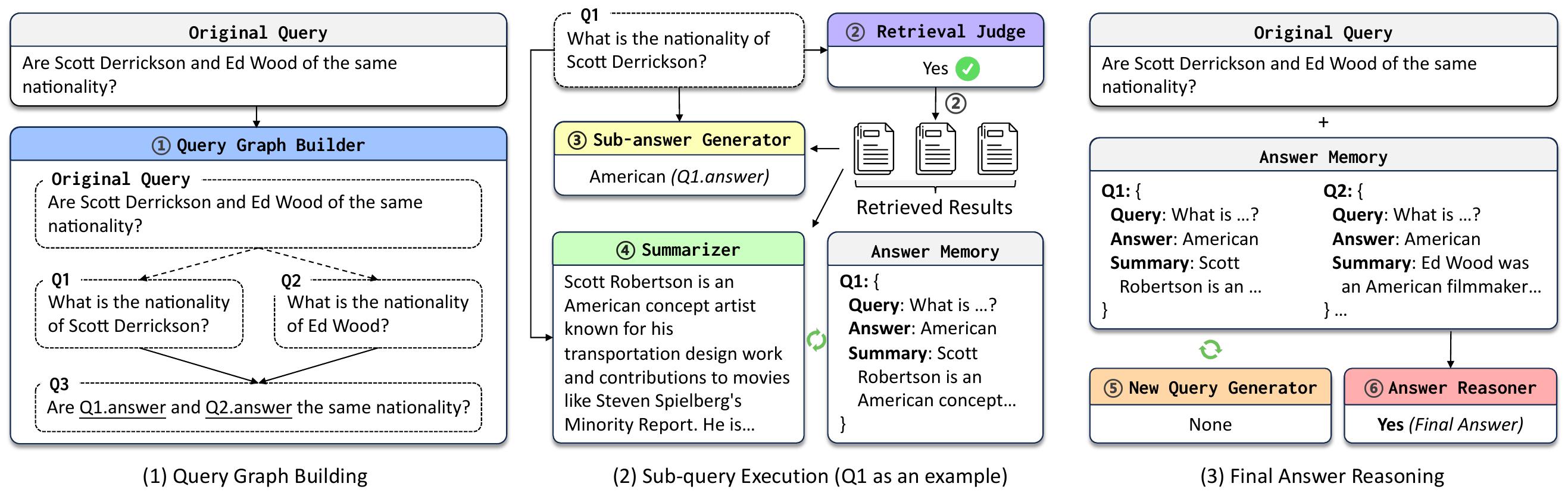}
    \caption{The illustration of our \ours{} framework, which contains three main steps: (1) query graph building, (2) sub-query execution, and (3) final answer reasoning.}
    \label{fig:framework}
    \vspace{-12px}
\end{figure*}

\section{Related Work}
\noindent\textbf{Retrieval-Augmented Generation}\quad
Retrieval-augmented generation (RAG) integrates a retrieval module that accesses external knowledge to enhance generation quality~\cite{DBLP:conf/nips/LewisPPPKGKLYR020,replug,flare,ircot,itergen}. 
Efforts to improve RAG systems have been made in different aspects. For example, some studies have aimed to improve query understanding, thus improving retrieval accuracy and overall generation quality~\cite{rqrag,planxrag}. Others have investigated the necessity of retrieval to minimize unnecessary retrieval calls, which in turn improves system efficiency and reduces the impact of irrelevant knowledge~\cite{slimplm,selfrag}. Additionally, refining retrieval results has been explored to mitigate the need for processing extensive input lengths and to reduce retrieval-related noise~\cite{llmlingua,recomp}. In a different vein, some studies have tried to advance the RAG pipeline, exploring strategies such as enabling LLMs themselves to determine when retrieval is beneficial during generation~\cite{flare}, interleaving retrieval with chain-of-thought reasoning~\cite{ircot}, and synergizing retrieval and generation in an iterative manner~\cite{itergen}. Our study improves the RAG pipeline by dividing sub-tasks within RAG into different modules and employing role-specific token optimization to selectively activate various LLM capabilities using designated role tokens. As all modules are carefully fine-tuned, the overall performance of our framework can be effectively improved.

\noindent\textbf{Query Understanding and Decomposition}\quad
Query understanding aims at inferring the intent of a user query, which is a critical component in a retrieval system~\cite{DBLP:journals/jiis/ArensKS96,DBLP:journals/ipm/AzadD19,DBLP:journals/corr/abs-2305-14283}. It involves a series of techniques such as query classification, expansion, rewriting, and suggestion. Recent studies have indicated that query understanding is also very important in RAG systems~\cite{DBLP:journals/corr/abs-2305-14283,DBLP:conf/emnlp/MaoJCL00X0C024}, as it determines the integration of external knowledge into the generation process. Among these studies, query decomposition has been particularly effective for handling complex queries by dividing them into more manageable sub-queries, thereby enhancing the accuracy of the retrieval process~\cite{mindsearch,rqrag,ircot}. However, these methods either rely on large-sized API-based models for query decomposition or unable to adjust the query plan dynamically. In contrast, our \ours{} introduces a dynamic query graph, where each sub-query can be adjusted dynamically according to the system's memory state. Besides, \ours{} is built entirely on open-source LLMs, which significantly enhances its practical applicability and efficacy.

\begin{table*}[t]
    \centering
    \small
    \caption{The input and output of each module in our framework. The content in brackets depends on the result of retrieval judgment.}
    \vspace{-5px}
    \begin{tabular}{lllr}
    \toprule
        \textbf{Module} & \textbf{Input} & \textbf{Output} & \textbf{\# Training samples} \\
    \midrule
        Query Graph Builder & Original query & Query graph & 25,654 \\
        Retrieval Judge & Sub-query, answer memory & ``Yes'' / ``No'' & 65,823 \\
        Sub-answer Generator & Sub-query, (retrieved result) & Answer & 65,823 \\
        Summarizer & Sub-query, (retrieved result) & Summary & 53,002 \\
        New Query Generator & Original query, answer memory & New query / ``None'' & 25,654 \\
        Answer Reasoner & Original query, answer memory & Final answer & 25,654 \\
    \bottomrule
    \end{tabular}
    \label{tab:data}
    \vspace{-12px}
\end{table*}

\section{Methodology}
In this work, we introduce a unified RAG framework \ours{}, which has two characteristics: (1)~We propose a role-specific token optimization strategy, in which the LLM's parameters are frozen while only the embeddings of added role tokens are tuned. By this means, all modules in our framework can share the same base LLM and address specific tasks by integrating role tokens into the input. 
(2)~We design a query graph builder that decomposes a user query into multiple sub-queries and organizes them as a directed acyclic graph (DAG). By dynamically resolving each sub-query node, the final answer can be more accurately generated. 

\subsection{Module Design}
As shown in Figure~\ref{fig:framework}, \ours{} divides the RAG process into three stage: (1) query graph building, (2) sub-query execution, and (3) final answer reasoning. In the first stage, a \textit{Query Graph Builder} decomposes the user input query $Q$ into $n$ sub-queries $\{q_i\}_{i=1}^{n}$, forming a DAG $G(Q)$. Then, in the second stage, for each sub-query $q_i\in G(Q)$, a \textit{Retrieval Judge} evaluates whether it can be directly answered by the LLM; if not, it will call a retriever to get relevant knowledge. An \textit{Answer Generator} then produces an answer $a_i$ for $q_i$, utilizing the retrieved knowledge if required. Simultaneously, a \textit{Summarizer} condenses the retrieved knowledge relevant to the generated answer. Each resolved sub-query, along with its answer and summary, is stored in an answer memory $M$. Upon resolving all sub-queries, the framework moves into the third stage, where a \textit{New Query Generator} examines $M$ and $Q$ to determine if additional knowledge is needed, potentially generating new queries handled in the same manner as the previous sub-queries. Finally, an \textit{Answer Reasoner} synthesizes the final answer $A$ from the accumulated data in the answer memory $M$. This systematic approach ensures comprehensive and accurate query resolution. Below, we introduce each module briefly.

\noindent\textbf{Query Graph Builder}\quad
The query graph builder constructs a representation of a user's reasoning plan as a DAG. As illustrated in the left side of Figure~\ref{fig:framework}, the original query is decomposed into two sub-queries (\ie, Q1 and Q2), and is ultimately represented by a final node (\ie, Q3). The dependency among sub-queries is described using a parent-child relationship, where a node becomes a child if its resolution relies on the answers from preceding nodes. Within each node, placeholders are utilized to denote the answers from parent nodes (\eg, Q1.answer and Q2.answer in Q3), which are substituted with actual values during execution process. This DAG structure ensures that the reasoning plan follows the Markov assumption, allowing for the resolution of the final node once all preceding nodes have been addressed. 

\noindent\textbf{Retrieval Judge}\quad
Previous studies have indicated that not all user queries require external retrieved knowledge, and in some cases, irrelevant knowledge may even hurt the LLM's performance~\cite{DBLP:conf/iclr/YoranWRB24,slimplm}. Therefore, we design a retrieval judge module to determine whether a sub-query can be directly resolved by the LLM. Only if the judgment result is ``False'', the retriever $\mathcal{R}$ will be called and provide relevant knowledge $K=\mathcal{R}(q_i)$. To improve the judgment accuracy, the LLM is provided with access to an answer memory (described later), which contains information from previous sub-queries.  This setup enables the retrieval judge to perform a \textit{removing} operation on the query graph, effectively minimizing unnecessary retriever activations and thereby enhancing the efficiency of the overall system.

\noindent\textbf{Sub-answer Generator}\quad
The sub-answer generator is tasked with producing responses based on the sub-queries and any associated retrieved knowledge. Due to the more focused nature of the sub-queries compared to the original query, the answers generated are typically more accurate. Upon generating a sub-answer, it is stored in an answer memory $M$, which is structured as a Python dictionary. Each sub-answer is keyed by its corresponding sub-query identifier for easy retrieval and reference. For instance, if the active sub-query is ``Q1'', both the sub-query content and its answer are stored in the dictionary $M$ under the key ``Q1''. This method ensures organized and efficient management of generated answers throughout the processing sequence.

\noindent\textbf{Summarizer}\quad
When the retriever is activated, the retrieved knowledge will facilitate the answer generation process. Since the sub-queries may be dependent, it is valuable to also store the retrieved knowledge for future sub-query resolution process. However, 
directly storing all retrieved knowledge is ineffective. Our observations indicate that the answer to a sub-query often serves as a bridge between related sub-queries. Therefore, we introduce a summarizer that condenses the retrieved knowledge $K$ while prioritizing the essential information about the answer.

\noindent\textbf{New Query Generator}\quad
Once all sub-queries have been answered, the final answer can be inferred from the answer memory. However, not all sub-queries can be ideally resolved. Therefore, we introduce a new query generator module that suggests additional sub-queries when necessary. This module takes the original query $Q$ and the answer memory $M$ as input, and outputs either a new sub-query $q_j$ or a termination signal (``None''). \revise{If a new query is generated, it will be added into the graph as a child of the final node and resolved by the sub-query execution process.} Functionally, this module performs an \textit{addition} operation on the query graph, enriching it with supplementary sub-queries to incorporate additional knowledge, thereby improving the completeness of the retrieval.

\noindent\textbf{Answer Reasoner}\quad
As the final step, the answer reasoner leverage the original query $Q$ and the updated answer memory $M$ to drive the final answer $A$. Since the answer memory retains all sub-queries, their corresponding answers, and relevant knowledge summary, the answer reasoner can synthesize such information to generate a well-grounded response to the original query.

\subsection{Data Collection}
Manually annotating data for training each component in \ours{} requires extensive human resources, which is impractical for our research. Consequently, we employ an expert LLM, specifically \texttt{Llama-3.1-70B-Instruct}, to generate training data automatically.
Specifically, we conduct the RAG process following our designed workflow, construct the input of each component using different prompts (provided in Appendix~\ref{app:prompt}), and record the corresponding output as raw data. By this means, we can automatically collect amounts of data without human intervention. 

However, the absence of golden annotations for evaluating the quality of each component's output poses a challenge. To address this, we borrow the idea of outcome reward models from reinforcement learning~\cite{DBLP:journals/corr/abs-2211-14275,DBLP:conf/naacl/YuGW24} and use the \textit{final answer quality} as a delegate for evaluation. Concretely, we compare the final answer ${A}$ generated by the expert model with the golden answer $\hat{A}$ provided by the dataset and compute the metric as $s = g(A, \hat{A})$.\footnote{The metrics can be exact matching score or F1 score that are commonly used in answer evaluation.} Data samples only where the final answer score $s$ exceeds a predetermined threshold $\alpha$ are retained. Recent studies~\cite{DBLP:journals/corr/abs-2211-14275} have shown that the outcome reward model can provide effective signals for model performance verification, so we believe our strategy can ensure the high-quality of the generated data. Table~\ref{tab:data} shows the statistics of our collected data.

\begin{figure}
    \centering
    \includegraphics[width=\linewidth]{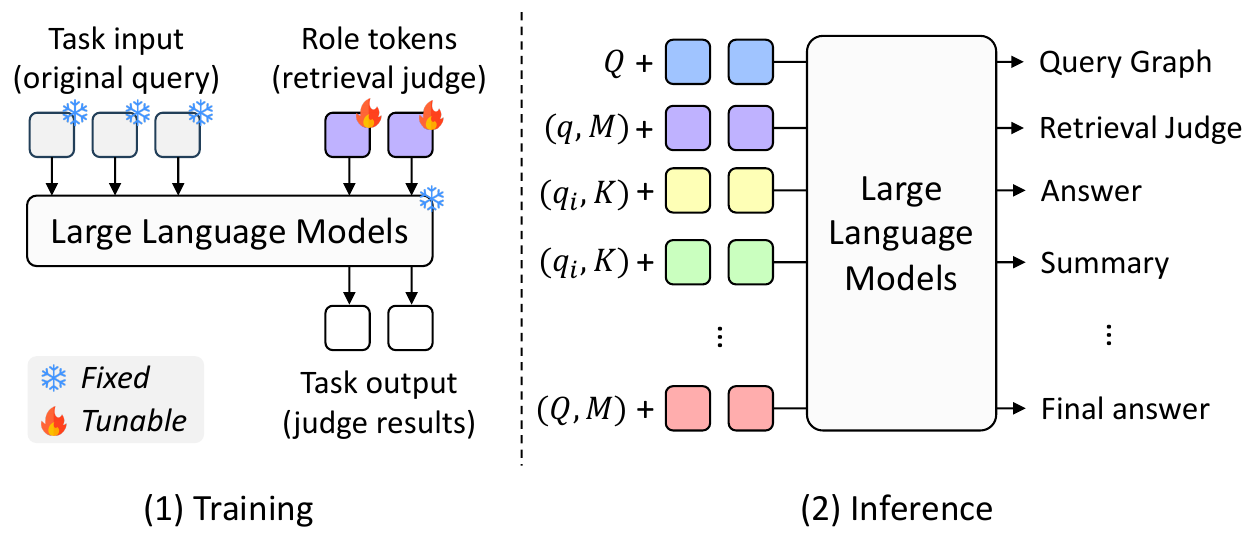}
    \caption{Illustration of the training and inference processes in our framework}
    \label{fig:mpt}
    \vspace{-12px}
\end{figure}

\subsection{Training Strategy}
\ours{} contains six interconnected modules, making it challenging for LLMs to learn and balance their abilities across different tasks. An ideal training strategy should meet three requirements: (1) It should be parameter-efficient as tuning LLMs is often expensive; (2) It should maintain the LLM's general ability as it may serve for different purposes in practice (including non-RAG scenarios); and (3) It should help the LLM understand various tasks in different components while facilitating seamless extension to new tasks.

To tackle these challenges, we propose a role-specific token optimization strategy, illustrated in Figure~\ref{fig:mpt}. The core idea is to use specialized role tokens to facilitate task-specific behavior in LLMs. \revise{We implement this by expanding the LLM's vocabulary with new special tokens designated for optimization, thereby preserving the integrity of the LLM's parameters.} During training, only these newly added tokens' embeddings are tuned, ensuring parameter efficiency while preserving the original LLM weights. This naturally satisfies the first two requirements. For the third requirement, since the added tokens are role-specific, they can be tailored for the task and not affect each other in training. Besides, it is easy to extend our framework with new modules by adding new role tokens. Specifically, for a specific task $T$ and a sample input $X^T$, we add several new tokens $[t_1,\ldots,t_n]$ and reformulate the input as $[X;t_1;\ldots;t_n]$, where $[;]$ is the concatenation operation. The next-token prediction objective can be defined as:
\begin{align}
    p = \prod_{i=1}^{m}p_{\theta,\delta}(y^T_i|X^T;\underbrace{t_1;\ldots;t_n}_{trainable};y^T_{<i}),
\end{align}
where $Y^T=[y^T_1,\ldots,y^T_m]$ is the target output, $\delta\in\mathbb{R}^{n\times d}$ represents the trainable parameters of the role tokens (\ie, their embeddings), and $d$ is the embedding size of the LLM. $\theta$ denotes the parameters of the backbone LLM, which are \emph{frozen} during training. Given that $|\delta| \ll |\theta|$, this method is \emph{highly efficient}. For example, with the Llama-3-8b model (where $d=4,096$), introducing $n=30$ tokens results in only 0.1M parameters. 

The inference stage is shown in the right side of Figure~\ref{fig:mpt}. \ours{} only deploys a single LLM, where task-specific role tokens are appended to the input to guide the LLM in performing different tasks effectively.

\section{Experiments}
\subsection{Datasets and Evaluation Metrics}
We conduct our experiments on five question-answering~(QA) datasets: HotpotQA~\cite{hotpotqa}, MuSiQue~\cite{musique}, 2WikimultihopQA~\cite{2wikimultihopqa}, Bamboogle~\cite{bamboogle}, and PopQA~\cite{popqa}. The details of these datasets are presented in Appendix~\ref{app:dataset}. We mix the training set of HotpotQA, MuSiQue, and 2WikimultihopQA for constructing our training set, and the remaining two datasets are used as head-out datasets. Among these datasets, PopQA~\cite{popqa} is the only dataset consisting merely of single-hop queries, which allows us to evaluate the generalizability of our approach to simpler queries. For evaluation, we primarily use the test sets provided by each dataset. If a test set is unavailable, we substitute it with the development set. Importantly, although some datasets have provided golden reference passages for the answer, we choose to use only the passages retrieved from the retrieval sets in both training and inference stages, which aligns with practical applications. Exact match (EM) and F1 score are employed as evaluation metrics.

\subsection{Baselines}
In addition to comparing with direct generation, we consider two kinds of RAG methods as baseline: 

(1) \textbf{Sequential pipeline}: These methods follow a standard \textit{retrieve-then-read} flow and focus on improving specific RAG components (\eg, query rewriting). \revise{Six representative methods are selected, including Standard RAG, SKR~\cite{skr}, SuRe~\cite{sure}, Trace~\cite{trace}, Adaptive-RAG~\cite{adaptiverag}, and BlendFilter~\cite{blendfilter}.}

(2) \textbf{Iterative pipeline}: This kind of method adjusts the sequential RAG pipeline by involving multiple cycles of retrieval and generation to refine outputs iteratively. \revise{We select five typical methods as baselines, including Self-RAG~\cite{selfrag}, IRCoT~\cite{ircot}, Iter-Retgen~\cite{itergen}, RetRobust~\cite{retrobust}, and RQ-RAG~\cite{rqrag}.}

Notably, some recent API-based models~\cite{planxrag,autorag} are not selected as baselines, because they do not integrate seamlessly with open-source LLMs. We provide an anonymous link of our code for review,\footnote{\url{https://anonymous.4open.science/r/RoleRAG}} and the implementation details are provided in Appendix~\ref{app:imple}.

\begin{table*}[t]
    \centering
    \small
    \caption{Experimental results of all methods using \texttt{LLaMA-3.1-8b} as the backbone model. The left three datasets are used for training \ours{}, representing in-domain evaluation, while the right two datasets are used for out-of-domain evaluation. The best and second-best results are highlighted in \textbf{bold} and \underline{underlined}, respectively.}
    \vspace{-5px}
    \begin{tabular}{lcccccccccc}
    \toprule
    & \multicolumn{2}{c}{HotpotQA} & \multicolumn{2}{c}{MuSiQue} & \multicolumn{2}{c}{2WikiMultihopQA} & \multicolumn{2}{c}{Bamboogle} & \multicolumn{2}{c}{PopQA} \\
    \cmidrule(r){2-3}\cmidrule(lr){4-5}\cmidrule(lr){6-7}\cmidrule(lr){8-9}\cmidrule(l){10-11}
    Method & EM & F1 & EM & F1 & EM & F1 & EM & F1 & EM & F1 \\ 
    \midrule
    Direct Generation & 16.20 & 25.15 & 3.30 & 9.30 & 16.50 & 26.30 & 9.60 & 16.13 & 11.10 & 20.65  \\
    \midrule
    \textit{Sequential pipeline} \\
    Standard RAG & 29.50 & 40.00 & 4.30 & 10.28 & 15.20 & 25.40 & 18.40 & 24.55 & 25.80 & 41.34 \\
    SKR & 24.20 & 34.85 & 3.40 & 9.67 & 15.70 & 26.50 & 12.80 & 19.43 & 19.40 & 32.04 \\
    SuRe & 23.80 & 36.24 & 5.20 & 10.05 & 10.20 & 18.00 & 16.80 & 25.96 & 27.60 & \underline{44.94} \\
    Trace & 26.00 & 35.30 & 5.60 & 11.30 & 9.50 & 15.80 & 13.60 & 19.60 & 26.60 & 39.29 \\
    Adaptive-RAG & 31.70 & {43.45} & 9.50 & 15.57 & 25.20 & 36.40 & 25.60 & 35.39 & 26.10 & 36.14 \\
    \revise{BlendFilter} & \underline{34.90} & \underline{45.56} & 7.70 & 13.54 & 24.30 & 33.19 & 22.40 & 31.04 & 25.40 & 41.01 \\
    \midrule
    \textit{Iterative pipeline} \\
    Self-RAG & 9.00 & 18.46 & 0.90 & 4.80 & 3.70 & 17.32 & 4.00 & 9.07 & 6.50 & 16.75 \\
    IRCoT & 30.50 & 40.62 & 9.70 & 15.42 & 27.60 & 36.20 & {30.40} & \underline{41.10} & {29.00} & 35.64 \\
    Iter-Retgen & {32.00} & 42.43 & 6.50 & 12.34 & 16.80 & 27.14 & 20.00 & 26.84 & 26.50 & 40.87 \\
    \revise{RetRobust} & 27.20 & 30.10 & 12.10 & 14.70 & \underline{32.20} & 33.50 & \underline{32.80} & 36.00 & \underline{32.80} & 37.00 \\
    RQ-RAG & 26.30 & 33.94 & \underline{10.20} & \underline{16.04} & {28.70} & \underline{37.64} & 24.80 & 32.18 & 15.60 & 31.37  \\
    \ours{} (ours) & \textbf{37.40} & \textbf{49.17} & \textbf{18.20} & \textbf{27.30} & \textbf{47.00} & \textbf{53.87} & \textbf{44.00} & \textbf{54.47} & \textbf{33.70} & \textbf{45.42} \\
    \bottomrule
    \end{tabular}
    \label{tab:main_result}
    \vspace{-10px}
\end{table*}

\subsection{Experimental Results}
The experimental results are shown in Table~\ref{tab:main_result}. It is evident to see that our \ours{} significantly outperforms other baseline methods on all five datasets. This clearly demonstrates the superiority of our method. We have further observations as:

(1) RAG methods generally outperform direct generation by a large margin, highlighting the advantage of integrating external knowledge for knowledge-intensive tasks. Specifically, iterative pipeline methods perform better than sequential pipeline methods. This is particularly evident in scenarios involving multi-hop queries, where the complexity often hinders the retriever's ability to gather all relevant information, leading to suboptimal generation performance. (2) Our \ours{} achieves the best performance in both in-domain and out-of-domain evaluations. This indicates that our proposed dynamic query graphs and multi-task prompt tuning effectively enhance RAG performance and exhibit strong generalizability. (3) On the single-hop QA dataset (PopQA), some iterative pipeline methods (\eg, RQ-RAG) underperform compared to sequential pipelines. This can be potentially attributed to the overly complex processing applied to relatively simple queries, which introduces unnecessary noise. In contrast, \ours{} can construct graphs with fewer nodes to represent simpler queries, which is more accurate and efficient. (4) We notice a poor performance of Self-RAG on several datasets, which has also been reported by recent studies~\cite{raglab}. By carefully checking its output, we find that Self-RAG tends to generate long reasoning paths that eventually mislead itself to generate incorrect answers. This may stem from its training strategy, which integrates all modules into a single generation process. Conversely, \ours{} employs independent training for each module using role tokens, which clarifies and simplifies the tasks each module must learn, thereby improving overall performance. 

\subsection{Further Analysis}

\begin{table}[t]
    \centering
    \small
    \caption{Performance (F1 score) of \ours{} with specific components removed.}
    \vspace{-5px}
    \setlength{\tabcolsep}{1.2mm}{
    \begin{tabular}{clcccc}
    \toprule
         & & \multicolumn{2}{c}{\textbf{HotpotQA}} & \multicolumn{2}{c}{\textbf{MuSiQue}}  \\
         \cmidrule(r){3-4}\cmidrule(l){5-6}
        \# & \textbf{Variant} & EM & F1 & EM & F1 \\ 
        \midrule
        1 & Full & 37.40 & 49.17 & 18.20 & 27.30 \\
        2 & $\hookrightarrow$ \revise{\textit{w/o} Q. Graph} & 31.30 & 42.60 & 5.80 & 12.54 \\
        3 & $\hookrightarrow$ \revise{\textit{w} Decompose prompt} & 29.40 & 39.98 & 6.90 & 13.31 \\
        4 & $\hookrightarrow$ \textit{w/o} Retrieval judge & 38.40 & 50.31 & 18.50 & 27.58  \\
        & \quad \textit{Save retrieval} & \multicolumn{2}{c}{\textit{22.56\%}}  & \multicolumn{2}{c}{\textit{9.2\%}}  \\
        5 & $\hookrightarrow$ \revise{\textit{w/o} Summarizer} & 31.20 & 42.47 & 5.70 & 12.44 \\
        6 & $\hookrightarrow$ \textit{w/o} New Q. gen & 37.20 & 49.07 & 18.00 & 27.30  \\
        & \quad \textit{Need new query} & \multicolumn{2}{c}{\textit{5.9\%}} & \multicolumn{2}{c}{\textit{14.10\%}}  \\
    \bottomrule
    \end{tabular}
    }
    \label{tab:ablation}
    \vspace{-10px}
\end{table}

\paragraph{Ablation Study}
\revise{We conduct comprehensive experiments to explore the contribution of each module in our framework, with results shown in Table~\ref{tab:ablation}. Our analysis first focuses on the query graph builder by examining two variants: complete removal  (\#2) and replacement with a prompt-based query decomposition approach (\#3). The results indicate that query decomposition is crucial for handling complex queries, and that LLMs struggle to perform this task effectively through direct prompting, highlighting the significance of our graph-based approach. The retrieval judge component demonstrates an interesting trade-off: while it causes a marginal decrease in performance due to reduced knowledge incorporation (\#4), it substantially reduces retrieval costs, thereby improving system efficiency. To evaluate the summarizer's impact, we implement a variant that simply uses the first retrieved passage for length control. The observed performance degradation confirms the important role of the summarizer. Finally, the new query generator improves performance by introducing additional useful knowledge, despite being activated in fewer than 15\% of queries, highlighting its effectiveness.}

\begin{table}[t]
    \centering
    \small
    \caption{Performance (F1 score) of \ours{} on two datasets using different LLMs.}
    \vspace{-5px}
    \setlength{\tabcolsep}{1.2mm}{
    \begin{tabular}{lllcc}
    \toprule
        \textbf{\#} & \textbf{Query graph} & \textbf{Other modules} & \textbf{HotpotQA} & \textbf{MuSiQue} \\
    \midrule
    \multicolumn{5}{c}{\ours{} \textit{Default setting}} \\
        1 & Llama-8B & Llama-8B & 49.17 & 27.30 \\
    \midrule
    \multicolumn{5}{c}{\textit{Using various LLMs as backbones}} \\
        2 & Llama-3B & Llama-3B & 41.50 & 20.94 \\
        3 & Mistral-7B & Mistral-7B & 47.62 & 25.51 \\
        4 & Llama-70B & Llama-70B & 53.46 & 29.86 \\
    \midrule
    \multicolumn{5}{c}{\textit{Using various LLMs for different modules}} \\
        5 & Llama-70B & Llama-8B & 50.65 & 27.53 \\
        6 & Llama-8B & Llama-70B & 54.23 & 28.19 \\
    \bottomrule
    \end{tabular}
    }
    \label{tab:model_size}
    \vspace{-10px}
\end{table}

\paragraph{Impact of Model Size}
The size of LLMs often determines their performance. Therefore, we investigate the impact of model sizes from two perspectives: (1) by using different backbone LLMs to drive the entire \ours{} framework, and (2) by replacing the core module (\ie, query graph builder) with various LLMs. We conduct experiments using \texttt{Llama-3.2-3B-instruct} (Llama-3B), \texttt{Llama-3.1-8B-instruct} (Llama-8B), \texttt{Llama-3.1-70B-instruct} (Llama-70B), and \texttt{Mistral-7B-Instruct-v0.3} (Mistral-7B). Notably, only the Llama-70B is applied using few-shot examples, while the others are fine-tuned on our training set. The results are shown in Table~\ref{tab:model_size}. First, we can observe that \ours{} consistently achieves promising results across all settings, demonstrating the method's robust versatility. Second, using larger LLMs generally leads to better performance (\#1-4). This is reasonable as larger models have strong abilities in language understanding and generation. Intriguingly, the Llama-70B model plays a better role in resolving queries (\#6) than in building the query graph (\#5). This suggests that while query decomposition is a complex task, it can be effectively learned with sufficient model training. Conversely, the ability to resolve queries appears to be more closely tied to the intrinsic performance of the model itself.


\begin{figure}[t]
    \centering
    \begin{subfigure}[b]{0.5\linewidth}
        \centering
        \includegraphics[width=\linewidth]{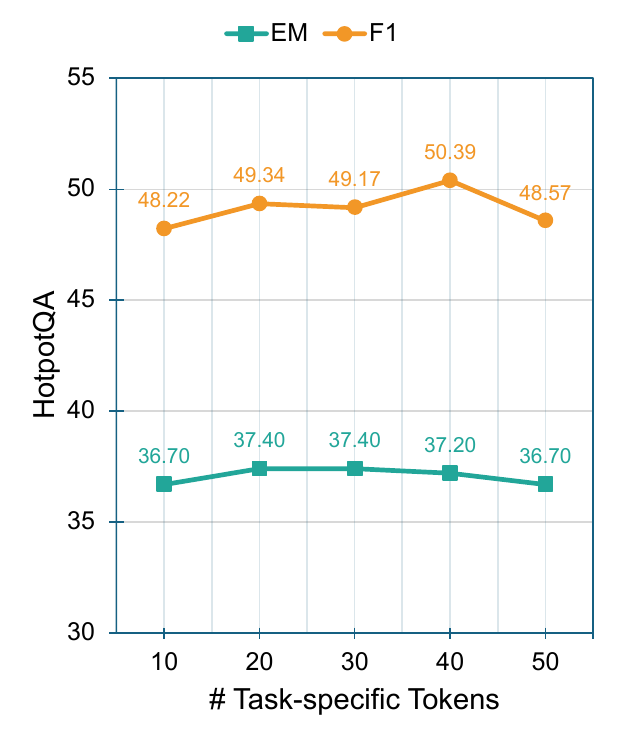}
    \end{subfigure}%
    \hfill
    \begin{subfigure}[b]{0.5\linewidth}
        \centering
        \includegraphics[width=\linewidth]{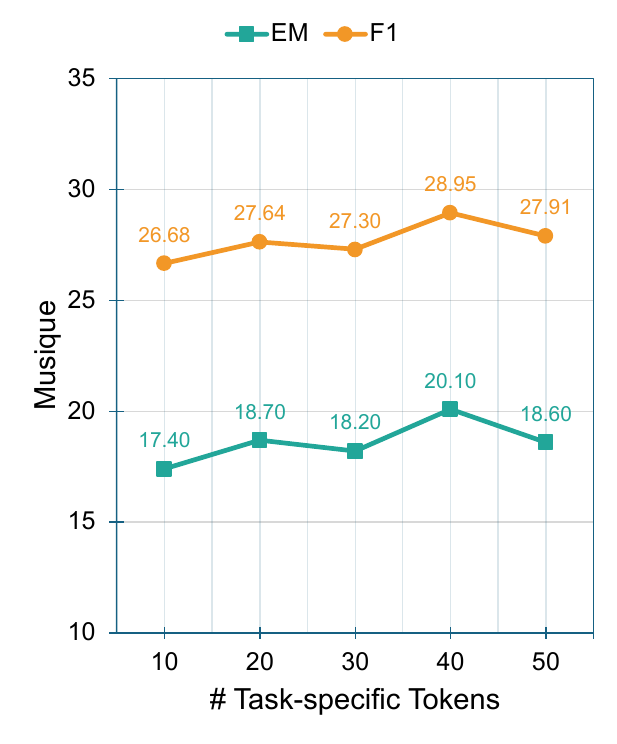}
    \end{subfigure}
    \caption{Performance with various numbers of tokens.}
    \label{fig:token_num}
    \vspace{-10px}
\end{figure}

\paragraph{Impact of Task-specific Token Amounts}
In \ours{}, we use task-specific tokens in multi-task prompt tuning to learn different tasks in RAG. We explore the correlation between the number of added tokens and the final performance, as shown in Figure~\ref{fig:token_num}. We can observe: (1) It is surprising that using only ten tokens per task can provide significant performance improvement, highlighting the efficiency of our approach. (2) The performance generally improves when more tokens are used, with optimal results occurring when 30-40 tokens are used per task (varies slightly across different datasets). Taking adding 30 tokens as an example, our method adds 0.86M parameters in total, which is only about 0.01\% of the full model, validating again its parameter efficiency. (3) However, further increasing the token amount does not improve performance; a decline is noted when 50 tokens are used per task, implying potential overfitting issues.

\begin{figure}
    \centering
    \includegraphics[width=\linewidth]{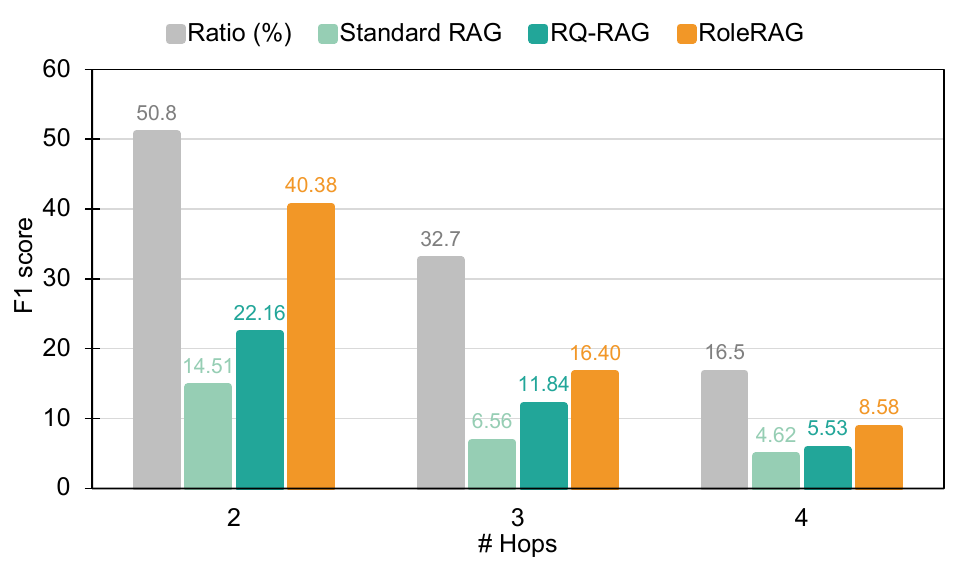}
    \caption{Performance of various models on questions of different complexity (MuSiQue). \revise{``Ratio'' indicates the proportion of a certain category to the entire data.}}
    \label{fig:hops}
    \vspace{-10px}
\end{figure}

\paragraph{Impact of Query Complexity}
An advantage of our framework lies in its ability to decompose complex queries into sub-queries and leverage multiple modules to resolve them. We employ the MuSiQue dataset, which contains human-annotated decomposition labels, to investigate the performance across query complexities. We analyze the performance of different models, including \ours{}, RQ-RAG, and Standard RAG, on questions categorized by the number of intermediate steps (hops) required. The experimental results are shown in Figure~\ref{fig:hops}. It is evident to see that the standard RAG method struggles with complex multi-hop queries, because the retriever cannot effectively gather comprehensive information that spans all facets of a query. In contrast, both RQ-RAG and \ours{} can iteratively resolve the sub-queries, which significantly improves the performance. Unfortunately, RQ-RAG learns both the query decomposing and query resolving tasks by a single model, making it challenging for the LLM to learn different abilities required by these tasks. Notably, our framework achieves over 60\% alignment with human-annotated decomposition results, while RQ-RAG reaches only 18\%. This highlights again the superiority of our \ours{}, which distinctively separates these tasks to optimize performance.

\begin{table}[t]
    \centering
    \small
    \caption{Performance (F1 score) of \ours{} with different numbers of passages per query (\# P. / Q.). \ours{}-$x$ denotes using $x$ passages per sub-query.}
    \vspace{-5px}
    \setlength{\tabcolsep}{1.7mm}{
    \begin{tabular}{llcccc}
    \toprule
        & & \multicolumn{2}{c}{\textbf{HotpotQA}} & \multicolumn{2}{c}{\textbf{MuSiQue}} \\
    \cmidrule(r){3-4} \cmidrule(l){5-6}
        \textbf{\#} & \textbf{Method} & \textbf{\# P. / Q.} & \textbf{F1} & \textbf{\# P. / Q.} & \textbf{F1} \\
    \midrule
        1 & Standard RAG & 5.00 & 40.00 & 5.00 & 10.28 \\
        2 & RQ-RAG & 7.35 & 33.94 & 7.43 & 16.04 \\
        3 & \ours{}-1 & 2.27 & 43.58 & 2.35 & 20.73 \\
        4 & \ours{}-2 & 4.54 & 46.73 & 4.70 & 25.08 \\
        5 & \ours{}-3 & 6.81 & 47.80 & 7.05 & 26.24 \\
        6 & \ours{}-4 & 9.08 & 48.66 & 9.40 & 27.07 \\
        7 & \ours{}-5 & 11.35 & 51.03 & 11.75 & 27.30 \\
    \bottomrule
    \end{tabular}
    }
    \label{tab:num_passages}
    \vspace{-10px}
\end{table}

\paragraph{Impact of Retrieval}
Since \ours{} decomposes original queries into multiple sub-queries, its superior performance may be benefited from more sufficient external knowledge. To examine this, we conduct experiments by adjusting the number of retrieved passages per sub-query, and the results are illustrated in Table~\ref{tab:num_passages}. We can observe that \ours{} can significantly outperform the standard RAG method, with fewer than half the retrieved passages (\#3 vs. \#1). This shows that query decomposition can indeed improve retrieval accuracy, which in turn enhances the overall performance of the RAG model. Compared with another iterative RAG pipeline RQ-RAG, \ours{} still has better performance, suggesting that it can construct sub-queries more accurately.

To provide a more intuitive understanding of our framework, we include a case study in Appendix~\ref{app:case}.

\section{Conclusion}
In this paper, we introduced \ours{}, a unified RAG framework that comprises six modules that collaborate to accomplish the full RAG process. To efficiently optimize these modules, we proposed a role-specific optimization strategy, which enhances the LLM's ability across diverse tasks by tuning only a small set of role tokens, while keeping the backbone model parameters frozen. Additionally, we structured the RAG process as a query graph resolution process, where dynamic sub-query resolution efficiently retrieves and supplements relevant knowledge. Through extensive experiments on multiple datasets, we demonstrated the effectiveness, generalizability, and flexibility of our method. 

\section*{Limitations}
This study introduces a unified RAG framework using role-specific token optimization. While our approach is highly effective, it has some limitations. First, the RAG process follows a predefined workflow, where all modules are activated in a fixed sequence. This restricts the framework's flexibility, as an ideal solution would allow the LLM to autonomously determine the workflow. Recent reinforcement learning methods could potentially enable such adaptive decision-making; however, collecting high-quality processing paths is challenging, and reinforcement learning itself is often unstable. Investigating an automatic workflow optimization remains an important direction for future work. Second, our framework processes each query by iteratively activating different modules, which may introduce efficiency overhead compared to directly feeding retrieved results and user queries into an LLM. Fortunately, when deployed as an online service, this efficiency issue can be mitigated. Since our framework leverages role tokens to modulate the LLM's functionality, it enables the batching of multiple LLM requests, significantly improving inference efficiency.


\clearpage
\appendix
\section{Discussion about Query Graph Builder}
\revise{In our framework, the query graph builder constructs a query graph before resolving each sub-query. Indeed, pre-constructing the query graph may constrain dynamic reasoning paths to some extent. However, during our preliminary experiments, we observe significant issues with purely iterative methods such as IRCoT~\cite{ircot} and Self-RAG~\cite{selfrag}, which tend to become overly reliant on intermediate retrieved results, thus propagating generation errors through subsequent reasoning steps. This may because current LLMs cannot accurately coordinate internal and external knowledge. As a result, we choose to build a query graph before resolving the query, and our experiments demonstrate that this strategy is effective for complex queries like the multi-hop questions in QA tasks. Additionally, our new query generator serves as a complementary iterative mechanism to dynamically enhance query exploration when necessary. We believe further exploring a hybrid approach that integrates planning-based and iterative methods is a promising direction for future work.}

\section{Prompt for Data Collection}\label{app:prompt}
\begin{figure*}
    \centering
    \includegraphics[width=\linewidth]{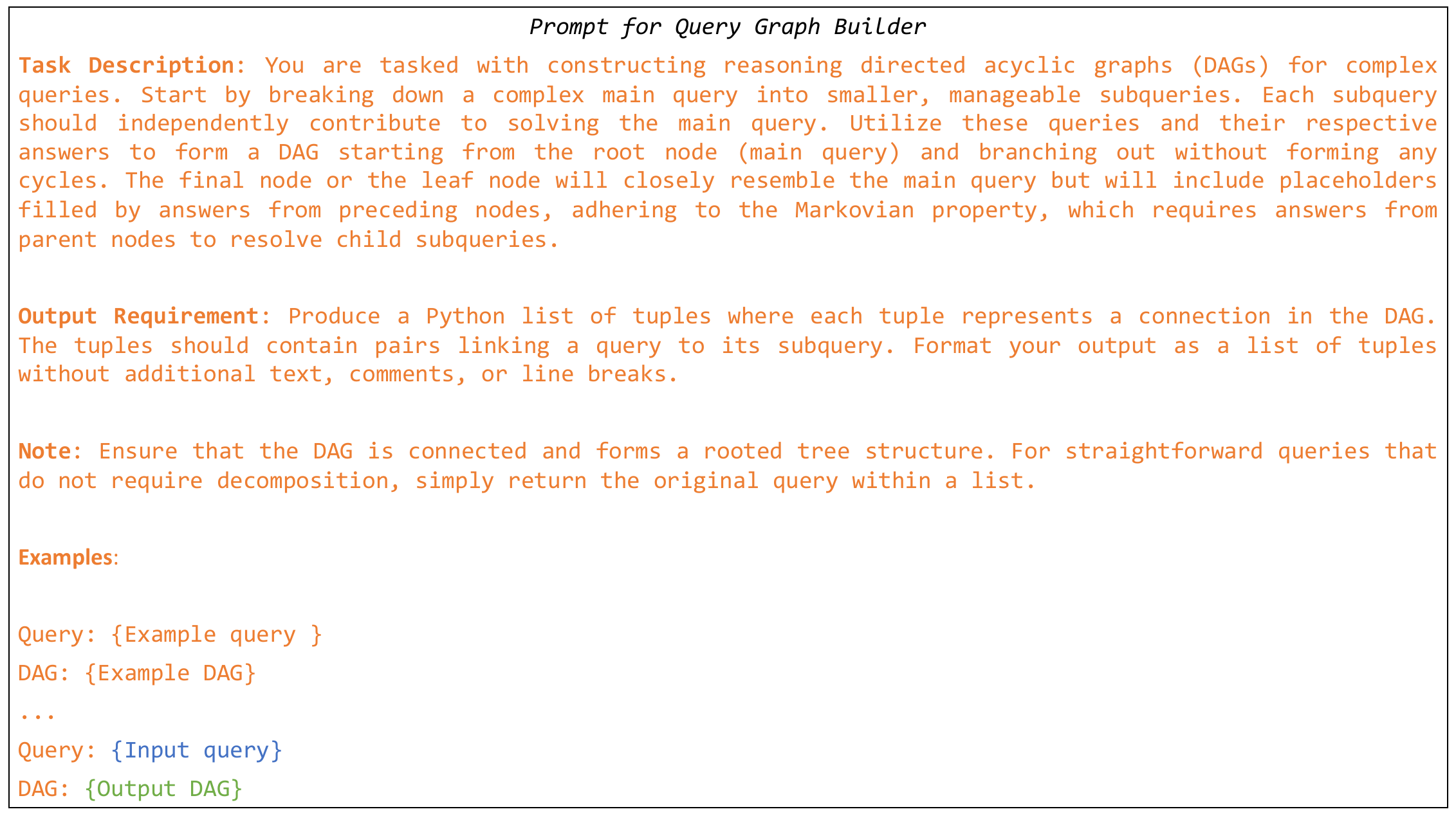}
    \caption{Prompt using for generating data for query graph builder.}
    \label{fig:p1}
\end{figure*}

\begin{figure*}
    \centering
    \includegraphics[width=\linewidth]{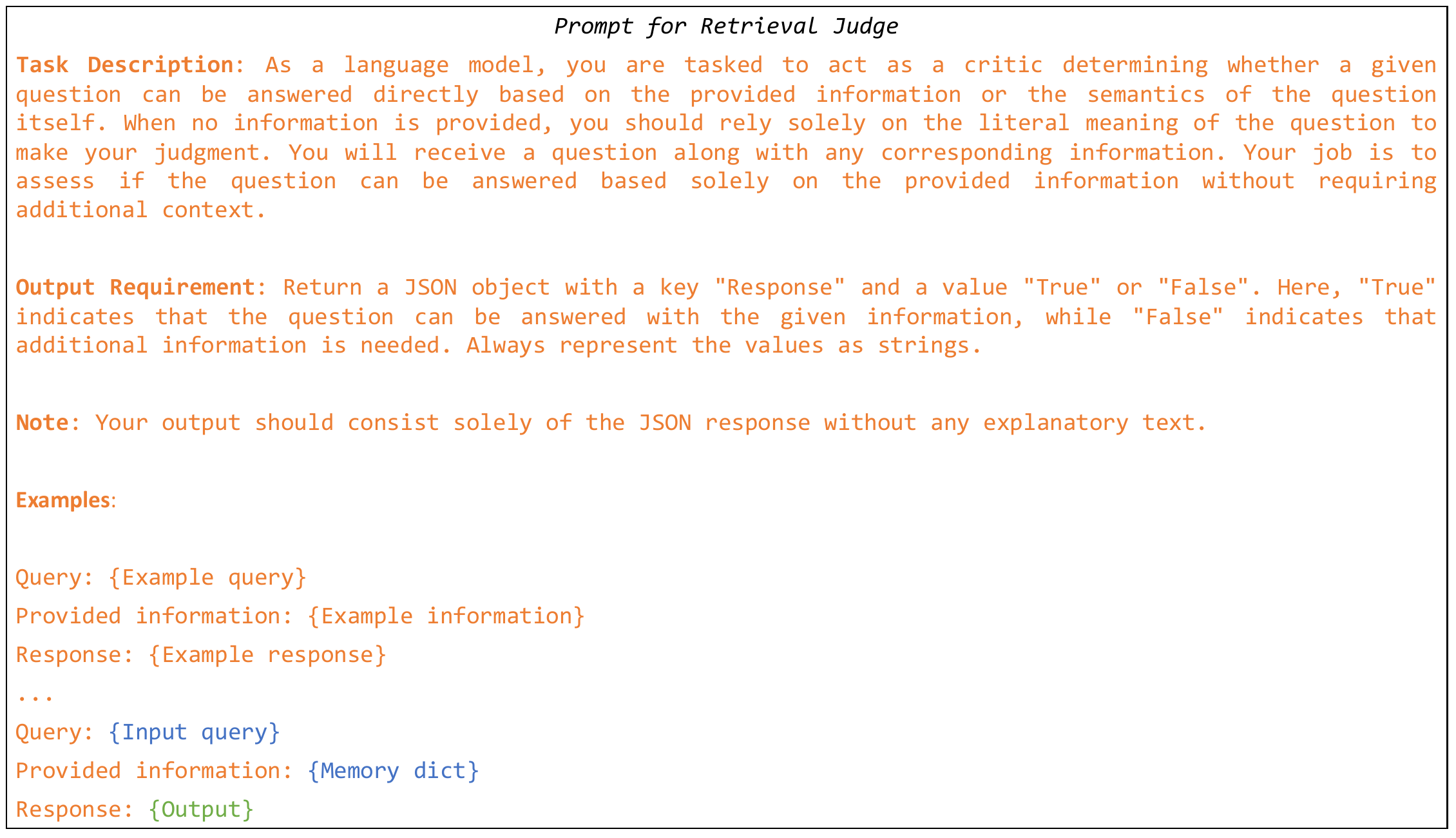}
    \caption{Prompt using for generating data for retrieval judge.}
    \label{fig:p2}
\end{figure*}

\begin{figure*}
    \centering
    \includegraphics[width=\linewidth]{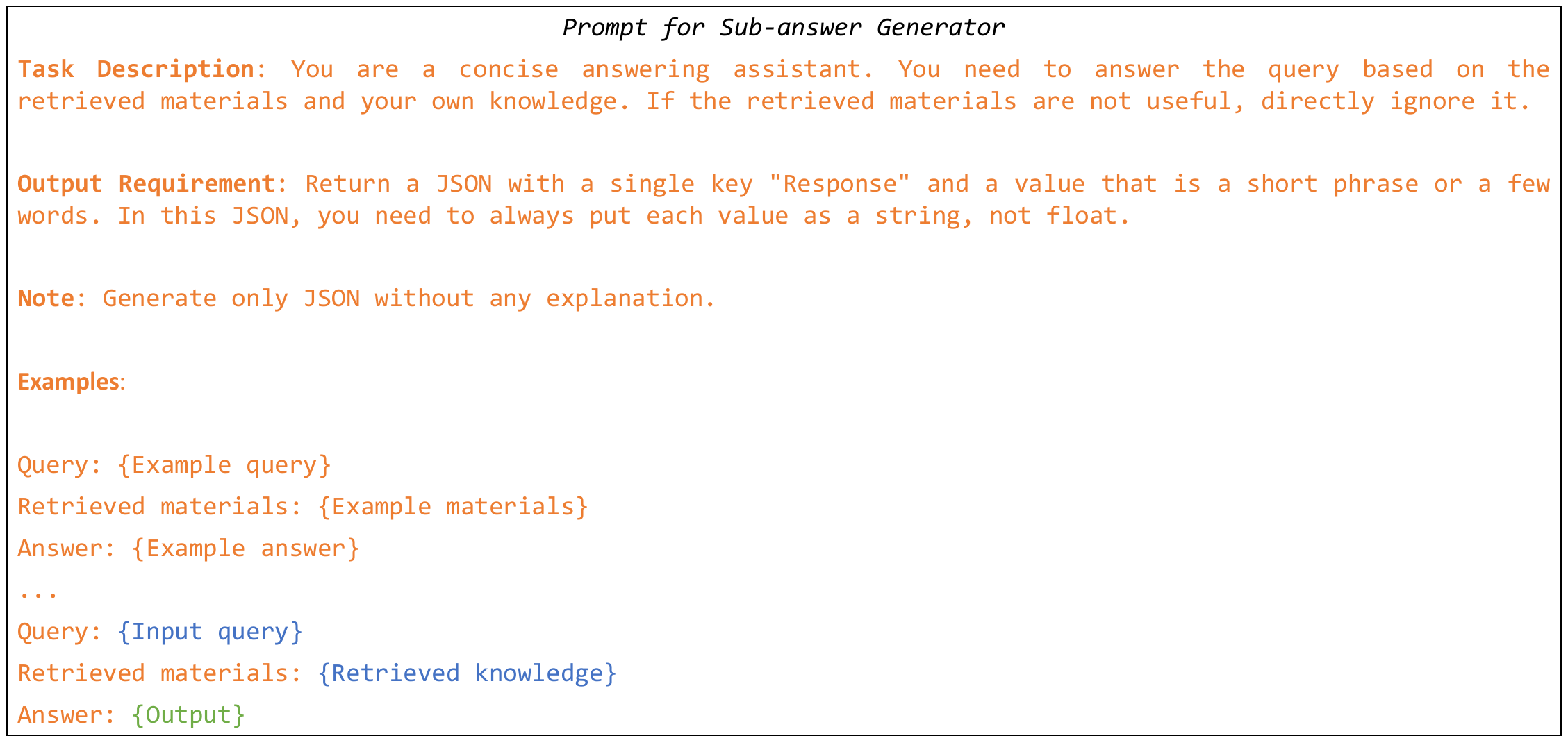}
    \caption{Prompt using for generating data for sub-answer generator.}
    \label{fig:p3}
\end{figure*}

\begin{figure*}
    \centering
    \includegraphics[width=\linewidth]{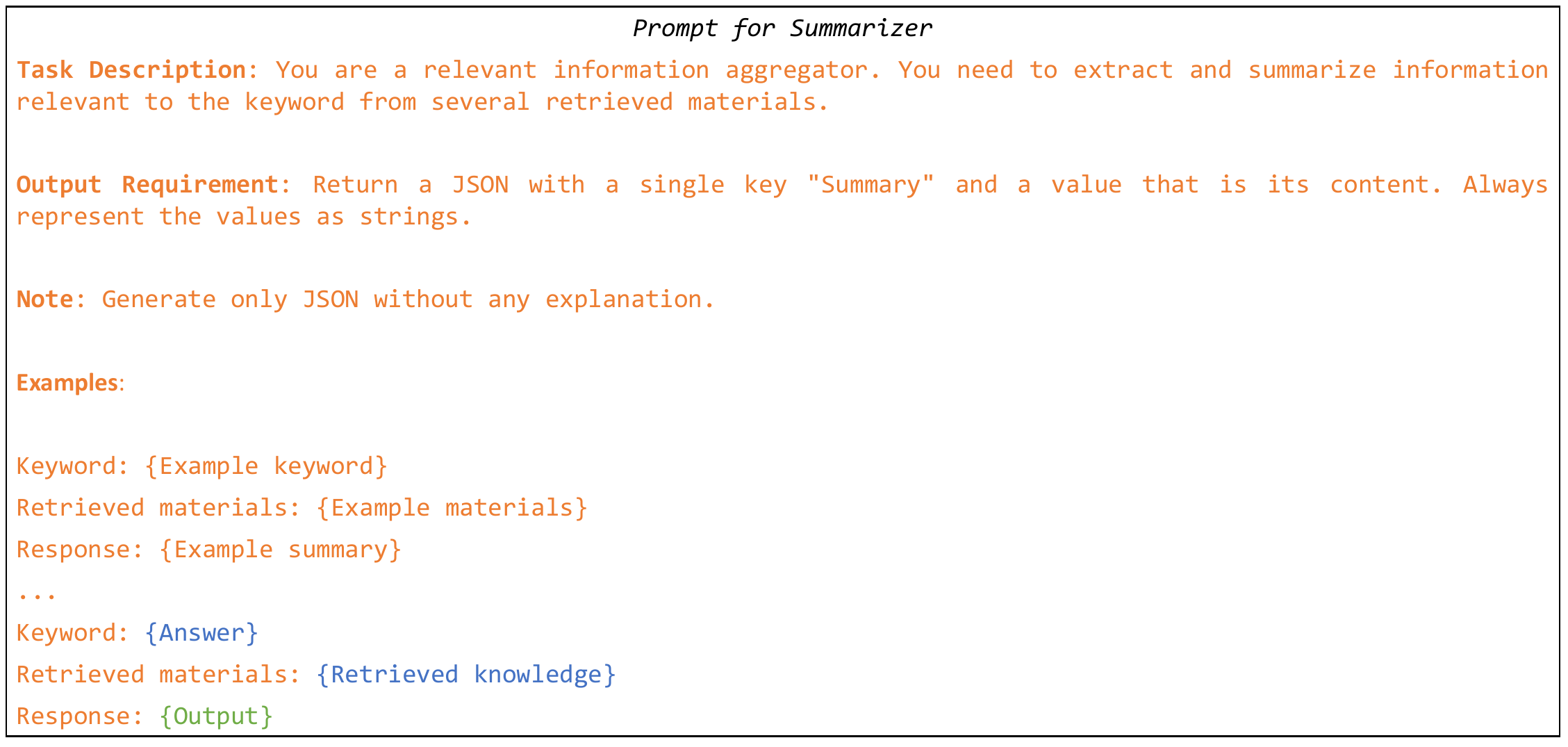}
    \caption{Prompt using for generating data for summarizer.}
    \label{fig:p4}
\end{figure*}

\begin{figure*}
    \centering
    \includegraphics[width=\linewidth]{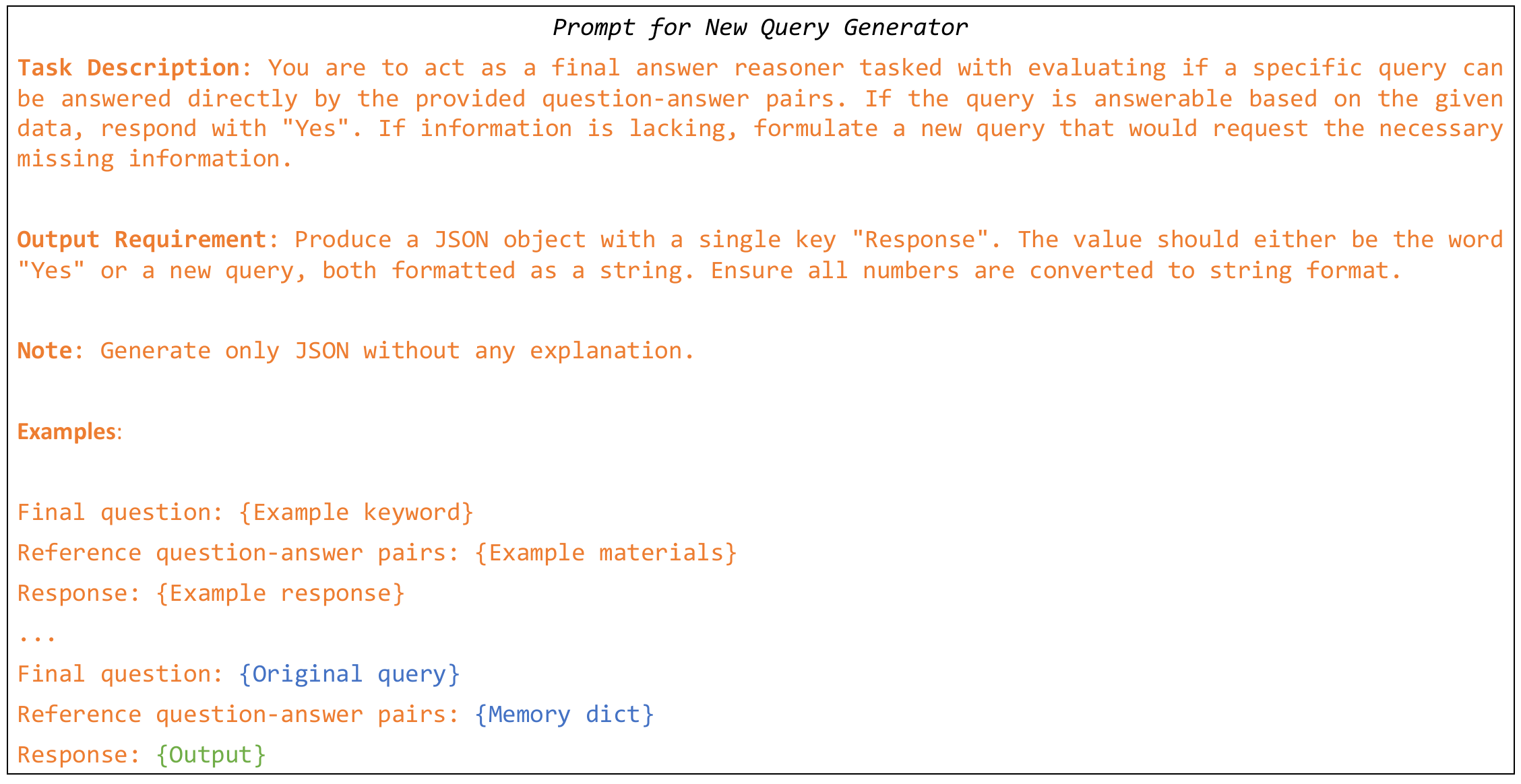}
    \caption{Prompt using for generating data for new query generator.}
    \label{fig:p5}
\end{figure*}

\begin{figure*}
    \centering
    \includegraphics[width=\linewidth]{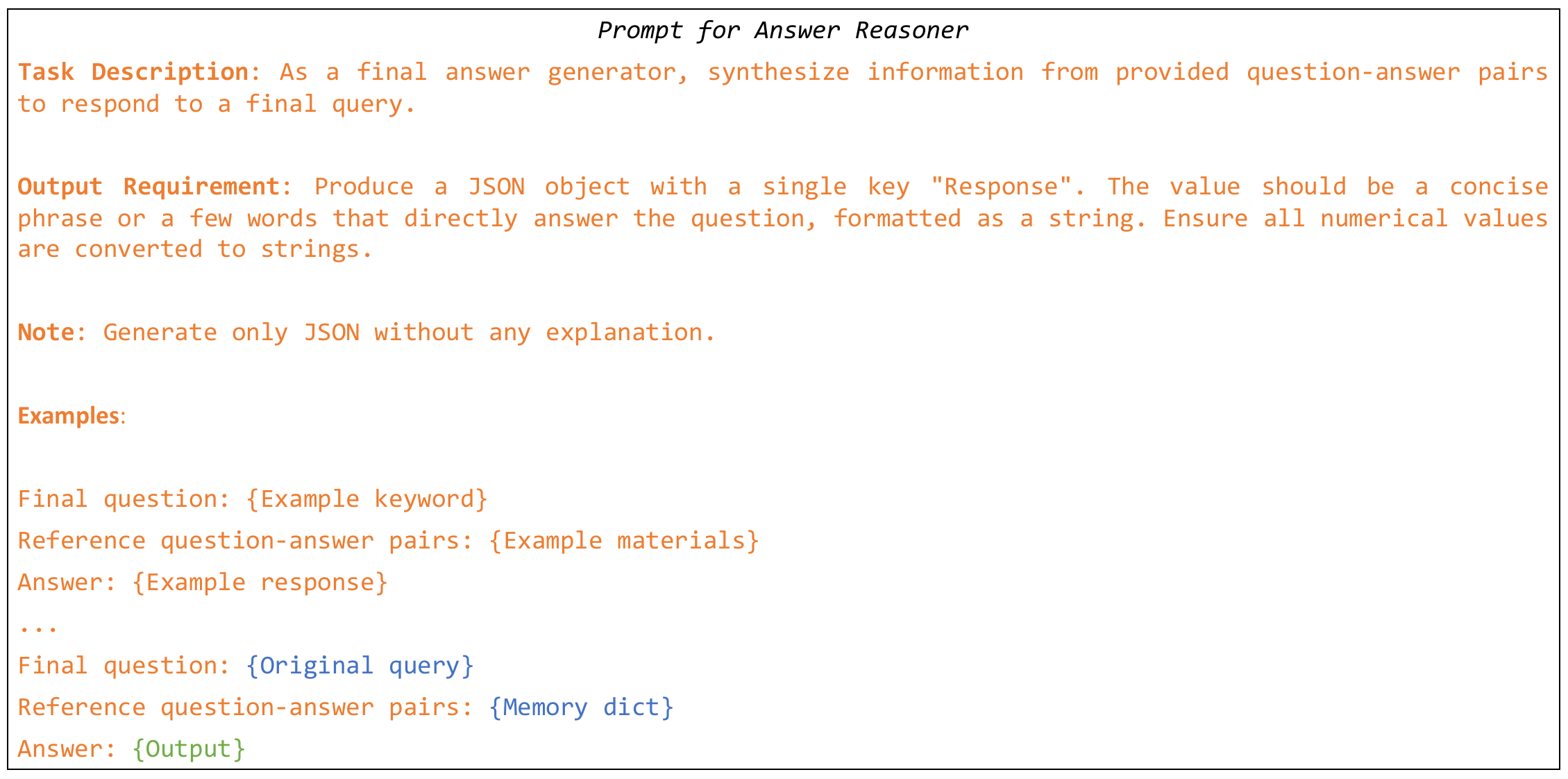}
    \caption{Prompt using for generating data for answer reasoner.}
    \label{fig:p6}
\end{figure*}

As illustrated in Figure~\ref{fig:p1}–\ref{fig:p6}, we manually craft different prompts and employ an expert LLM to generate training data. Each prompt consists of five key components: (1) Task description, providing context to help the LLM understand the task; (2) Output requirements, specifying the expected format and structure; (3) Guidelines, highlighting rules for data generation; (4) Demonstration examples, serving as in-context learning references; and (5) Task input, representing the specific instance to be processed.

\section{Details of Datasets}\label{app:dataset}
We conduct our experiments on five QA datasets, which are all provided by FlashRAG~\cite{flashrag} under the license of CC-BY-SA-4.0.\footnote{\url{https://creativecommons.org/licenses/by-sa/4.0/}}

\textbf{HotpotQA}~\cite{hotpotqa} is a large-scale QA dataset comprising Wikipedia-based question-answer pairs. Designed to facilitate complex reasoning, it features questions that require synthesizing information from multiple supporting documents. The dataset is diverse, unconstrained by pre-existing knowledge bases or schema. Additionally, HotpotQA introduces factoid comparison questions to assess a system's ability to extract and compare relevant information. 

\textbf{MuSiQue}~\cite{musique} is a multi-hop QA dataset designed to require genuine multi-hop reasoning. Each question necessitates 2 to 4 reasoning steps (hops). The dataset is constructed by systematically selecting and composing pairs of single-hop questions that are connected, ensuring that one reasoning step critically relies on information from another. This bottom-up methodology provides fine-grained control over the construction process and the properties of the resulting multi-hop questions. 

\textbf{2WikimultihopQA}~\cite{2wikimultihopqa} is a multi-hop QA dataset designed to evaluate complex reasoning across both structured and unstructured data. It comprises questions that require models to perform multiple reasoning steps, utilizing information from different Wikipedia articles. 

\textbf{Bamboogle}~\cite{bamboogle} is a curated dataset designed to assess the compositional reasoning abilities of language models. It comprises 125 questions that are intentionally challenging for standard search engines, like Google, to answer correctly. Each question requires the model to integrate information from multiple sources or perform multi-step reasoning to arrive at the correct answer. The dataset covers a wide range of topics and question formats.

\textbf{PopQA}~\cite{popqa} is a large-scale, open-domain QA dataset comprising entity-centric single-hop question-answer pairs. Each question is generated by converting a knowledge tuple from Wikidata into a natural language format using predefined templates. The dataset includes detailed annotations such as the subject entity, object entity, relationship type, and corresponding Wikidata identifiers. PopQA is designed to evaluate language models' abilities to recall factual knowledge, particularly focusing on less popular long-tail entities.


\section{Implementation Details}\label{app:imple}
We use PyTorch~\cite{pytorch} and Huggingface Accelerate library to implement our method. The learning rate is set as $5\text{e-}5$ with a warm-up ratio of $0.02$. Our method is trained for three epochs, with a training batch size of $32$. The maximum sequence length is set as 2,048 tokens. We use eight NVIDIA A800 GPUs for training. For the datasets and baseline methods, we use the version provided by FlashRAG~\cite{flashrag}, where \texttt{Llama-3.1-8B-instruct} is used as the default backbone LLM. For the retrieval sets, we follow previous studies~\cite{DBLP:conf/iclr/YoranWRB24} and use Wikipedia as the retrieval corpus. \texttt{E5-base-v2}~\cite{e5} is used as the retriever. 

\revise{
All of the methods in our experiments use the same retrieval corpus, retriever, and backbone LLM. Specifically: }

\revise{(1) Standard RAG, SKR, SuRe, Trace, BlenderFilter, IRCoT, and Iter-Retgen rely solely on prompt engineering strategies without additional model training.}

\revise{(2) Adpative-RAG involves training a query classifier on data sampled from SQuAD, NQ, TriviaQA, MuSiQue, HotpotQA, and 2WikiMultihopQA. We use the classifier provided by the original authors.}

\revise{(3) Self-RAG and RQ-RAG are trained using the same dataset as ours, utilizing the publicly available code provided by their authors.}

\begin{table*}[t]
\centering
\small
\begin{tabular}{p{.08\linewidth}p{.15\linewidth}p{.2\linewidth}p{.2\linewidth}p{.15\linewidth}p{.12\linewidth}}
\toprule
Method & Module & Input & Input tokens & Output & Output tokens \\ \midrule
RoleRAG & Query graph builder & Original query & $m$ & Sub-queries & $n*m$ \\ 
 & Retrieval judge & Sub-query & $n*m$ & Judge result (Yes or No) & $n$ \\ 
 & Sub-answer generator & Sub-query, retrieved passages & $n*m + n*k*l$ & (Sub-)answer & $n*t$ \\ 
 & Summarizer & Retrieved passages & $n*k*l$ & Summary & $n*l$ \\ 
 & New query generator & Answer memory & $n*(m+t+l)$ & New sub-query & $m$ \\ 
 & Answer reasoner & Answer memory & $n*(m+t+l)$ & Answer & $t$ \\ 
 & Total &  & $n*(4m+2kl+2t+2l)+m$ &  & $n*(m+t+l+1)+m+t$ \\ \midrule
IRCoT & Sub-query and Sub-answer generation & (previous) Sub-query, (previous) Sub-answer retrieved passages & $n*m + n*(k*l) + (n-1)*(k*l+t) + (n-2)*(k*l+t) ... + k*l+t$ & Sub-queries and Sub-answers & $n*(m+t)$ \\ 
 & Final answer generation & Original query, all retrieved passages & $m + n*k*l$ & Answer & $t$ \\ 
 & Total &  & $n*(m+\frac{n+3}{2}*k*l+\frac{n-1}{2}*t)$ &  & $n*(m+t)+t$ \\ \midrule
RQ-RAG & Sub-query generation & Original query & $m$ &  & $n*m$ \\ 
 & Answer generation & (Sub-)query, retrieved passages & $(n+1)*m + n*k*l + n*t$ & Answer & $n*t$ \\ 
 & Total &  & $n*(m+kl+t)+m$ &  & $n*(m+t)$ \\ 
\bottomrule
\end{tabular}
\end{table*}

\section{Efficiency Analysis}\label{app:efficiency}
\revise{
For sequential pipeline methods, they only conduct retrieval once, so their computational costs are lower but their performance is also relatively worse. For iterative pipeline methods (including ours), we theoretically analyze the computational costs of IRCoT, RQ-RAG, and our \ours{}. For clarity, we assume:}

\revise{(1) All queries/sub-queries have equal length $m$; answers/sub-answers have length $t$; each sub-query retrieves $k$ passages; and they have the same length $l$.}

\revise{(2) The summarizer in \ours{} produces a summary with the same length of a single passage, \ie, its length is also $l$.}

\revise{(3) Retrieval is assumed for each sub-query to simplify analysis.}

\revise{From the results, we can see IRCoT has the highest computational cost due to iterative processing and repeated input of all previous sub-results. Our \ours{} and RQ-RAG have a similar input token complexity ($O(nkl)$), but \ours{} produces additional output tokens due to the summarization step. Nevertheless, in practical scenarios, the retrieval is selective, thus reducing real-world overhead.}

\revise{In summary, while iterative methods naturally incur higher computational costs compared to sequential methods, our \ours{}'s additional costs remain moderate relative to its significantly improved performance. Furthermore, \ours{} is inherently parallelizable due to its modular design driven by role tokens, making it feasible for practical deployments.}

\section{Case Study}\label{app:case}
\begin{figure*}
    \centering
    \includegraphics[width=\linewidth]{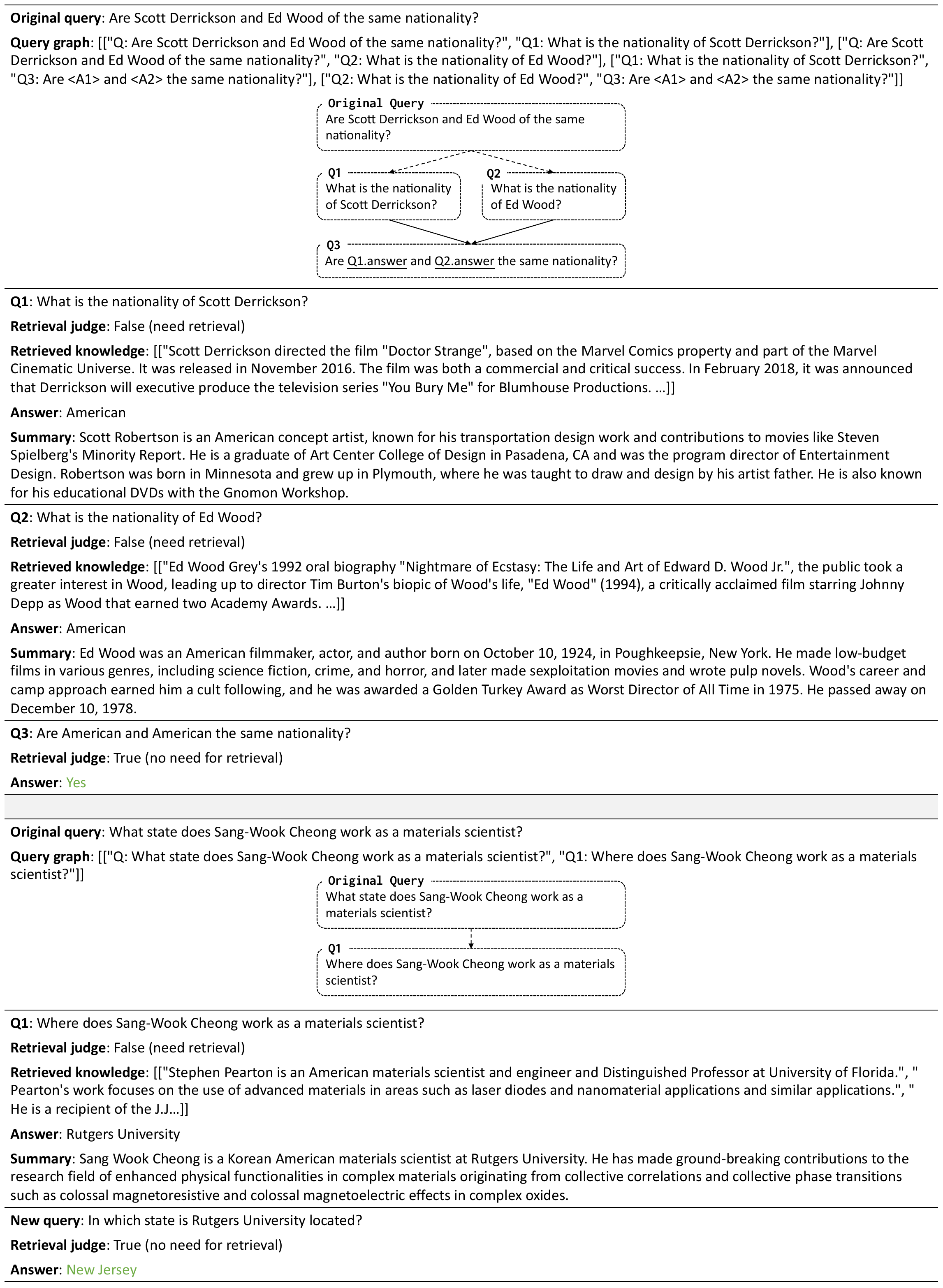}
    \caption{An example of \ours{} execution process from HotpotQA.}
    \label{fig:case1}
\end{figure*}

\begin{figure*}
    \centering
    \includegraphics[width=\linewidth]{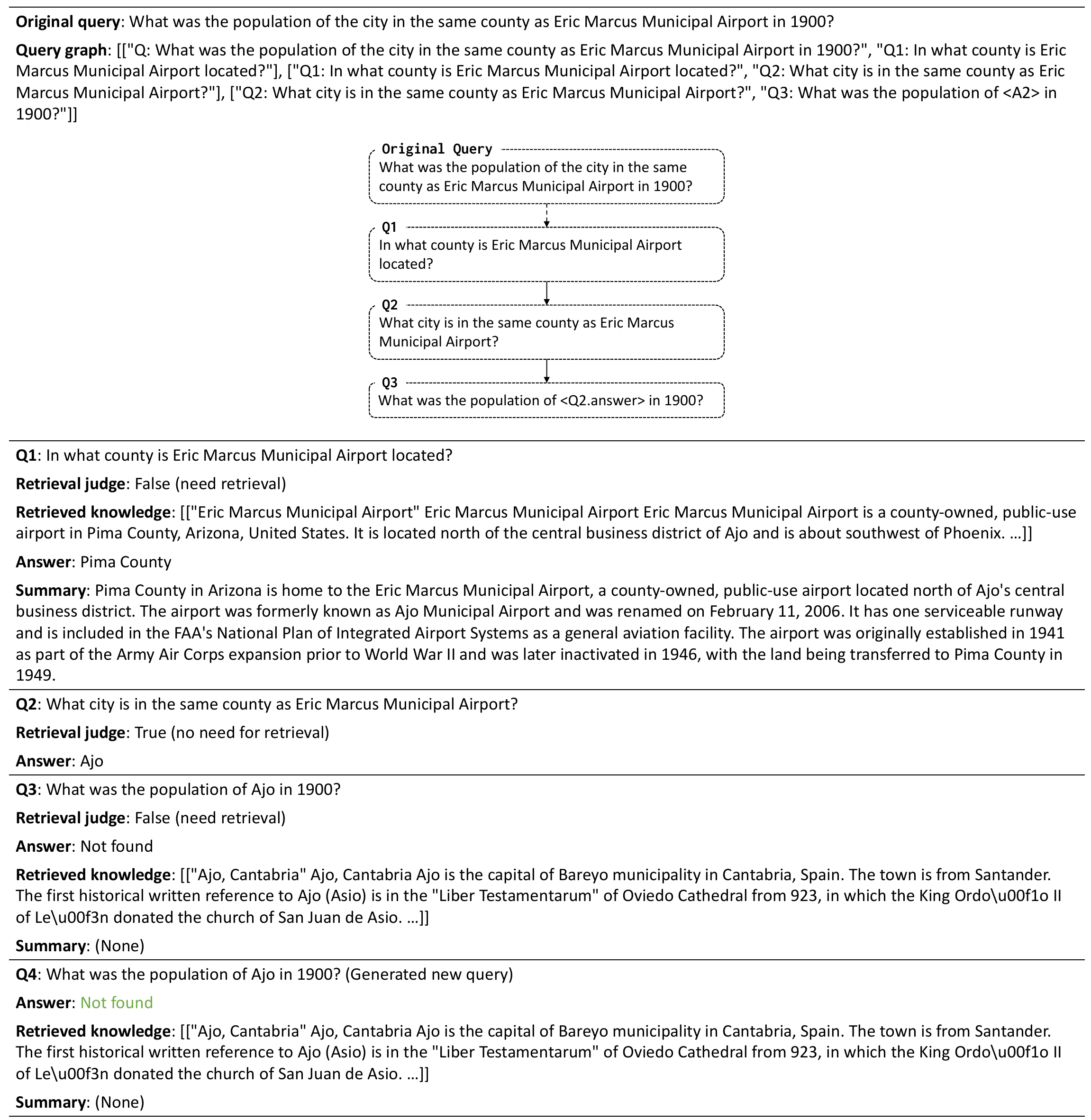}
    \caption{An example of \ours{} execution process from MuSiQue.}
    \label{fig:case2}
\end{figure*}

To further evaluate our framework qualitatively, we conduct a case study and present three representative examples in Table~\ref{fig:case1} and Table~\ref{fig:case2}. In the first case, our framework successfully decomposes the query into three sub-queries, where the third sub-query depends on the answers to the first two. By iteratively resolving the first two sub-queries, the answer to the third can be inferred directly without requiring additional retrieval. \revise{In the second case, our framework only rewrites the query, and the corresponding answer is incomplete. Fortunately, the new query generator successfully adds an effective sub-query to provide supplement information.} In contrast, the third case highlights a failure scenario. Although the original user query is split into three sequential sub-queries, the first two should be dependent, yet the model incorrectly treats them as independent. This suggests that accurately decomposing complex queries remains a challenging problem. Besides, while the second and third sub-queries are correctly formulated, the third sub-query fails to retrieve useful knowledge, leading to an incorrect final answer. In this scenario, the new query generator attempts to repeat the third sub-query. However, due to the limitations of the retrieval repository, the necessary information remains unavailable, resulting in an incorrect response. This case demonstrates that even when individual model components function correctly, external factors such as retrieval limitations can still prevent the system from generating the correct answer.

\begin{table}[t]
    \centering
    \small
    \begin{tabular}{lccc}
    \toprule
    & HotpotQA & MuSiQue & 2WikiQA \\
    \midrule
        Full & 49.17 & 27.30 & 53.87 \\
        \quad + Rewrite Ori. Q  & 50.04 & 27.19 & 53.49 \\
        \quad + Rewrite All  & 48.88 & 27.47 & 53.79 \\
    \bottomrule
    \end{tabular}
    \caption{Performance (F1 score) of \ours{} with query rewrite module.}
    \label{tab:add_rewrite}
\end{table}

\section{Impact of Query Rewrite}
Query rewriting~\cite{DBLP:conf/sigir/XuC96,DBLP:journals/csur/CarpinetoR12} addresses the problem of users' ambiguity and inaccurate queries by rewriting the user's original query, which is helpful in RAG systems~\cite{DBLP:conf/emnlp/MaoJCL00X0C024}. Recent studies have demonstrated that LLMs are capable of understanding user intents and providing more informational rewritten queries~\cite{DBLP:journals/corr/abs-2305-14283}. Motivated by these findings, we consider incorporating a query rewriting module in our framework and evaluate its impact under two settings: (1) applying query rewriting only to the original query and (2) applying it to all sub-queries. The experimental results are shown in Table~\ref{tab:add_rewrite}. Generally, we can observe that query rewriting does not consistently improve performance. When applied to the original query, it influences the query graph construction, leading to mixed results. Notably, improvements are observed only on the HotpotQA dataset. A closer inspection of the data reveals that HotpotQA queries are relatively well-formed, making query rewriting beneficial in this case. However, applying query rewriting to all sub-queries also yields unstable performance, likely because the sub-queries in our query graph are already simple and do not require further refinement. Given these findings, we exclude the query rewriting module from our final framework, as it introduces additional computational overhead without providing consistent benefits.

\end{document}